\documentclass[review]{elsarticle}

\usepackage{lineno,hyperref,subcaption}
\usepackage{algorithm}
\usepackage[noend]{algpseudocode}
\modulolinenumbers[5]

\journal{}

\begin{document}

\begin{frontmatter}

\title{A Deep Learning Based 6 Degree-of-Freedom Localization Method for Endoscopic Capsule Robots}


\author[address1]{Mehmet Turan}
\ead{turan@is.mpg.de}

\author[address2]{Yasin Almalioglu}
\ead{yasin.almalioglu@boun.edu.tr}

\author[address3]{Ender Konukoglu}
\ead{ender.konukoglu@vision.ee.ethz.ch}

\author[address1]{Metin Sitti}
\ead{sitti@is.mpg.de}


\address[address1]{Physical Intelligence Department, Max Planck Institute for Intelligent Systems, Stuttgart, Germany}
\address[address2]{Computer Engineering Department, Bogazici Univesity, Turkey}
\address[address3]{ Computer Vision Laboratory, Department of Information Technology and Electrical Engineering, ETH Zurich, Switzerland}

\begin{abstract}
We present a robust deep learning based 6 degrees-of-freedom (DoF) localization system for endoscopic capsule robots. Our system mainly focuses on localization of endoscopic capsule robots inside the GI tract using only visual information captured by a mono camera integrated to the robot. The proposed system is a 23-layer deep convolutional neural network (CNN) that is capable to estimate the pose of the robot in real time using a standard CPU. The dataset for the evaluation of the system was recorded inside a surgical human stomach model with realistic surface texture, softness, and surface liquid properties so that the pre-trained CNN architecture can be transferred confidently into a real endoscopic scenario. An average error of $7.1\%$ and $3.4\%$ for translation and rotation has been obtained, respectively. The results accomplished from the experiments demonstrate that a CNN pre-trained with raw 2D endoscopic images performs accurately inside the GI tract and is robust to various challenges posed by reflection distortions, lens imperfections, vignetting, noise, motion blur, low resolution, and lack of unique landmarks to track.
\end{abstract}


\end{frontmatter}


\section{Introduction}

In the past decade, the advances in material science enabled the fabrication of small, low cost devices in a variety of popular applications. Following these advances, untethered pill-size, swallowable capsule endoscopes with an on-board camera and wireless image transmission device have been developed and used in hospitals for screening the gastrointestinal tract and diagnosing diseases such as the inflammatory bowel disease, the ulcerative colitis and the colorectal cancer. Unlike standard endoscopy, endoscopic capsule robots are non-invasive, painless and more appropriate to be employed for long duration screening purposes. Moreover, they can access difficult body parts that were not possible to reach before with standard endoscopy (e.g., small intestines). Such advantages make pill-size capsule endoscopes a significant alternative screening method over standard endoscopy \citep{liao2010indications, nakamura2008capsule, pan2011swallowable, than2012review}.

However, current capsule endoscopes used in hospitals are passive devices controlled by peristaltic motions of the inner organs. The control over the capsule's position, orientation, and functions would give the doctor a more precise reachability of targeted body parts and more intuitive and correct diagnosis opportunity.  Therefore, several groups have recently proposed active, remotely controllable robotic capsule endoscope prototypes equipped with additional functionalities such as local drug delivery, biopsy and other medical functions \citep{goenka2014capsule, nakamura2008capsule, munoz2014review, carpi2011magnetically, keller2012method, mahoney2013managing, yim2014biopsy, petruska2013omnidirectional}. An active motion control is, on the other hand, heavily dependent on a precise and reliable real time pose estimation capability which makes the robot localization the key capability for a successful endoscopic capsule robot operation. In the last decade, several localization methods \citep{than2012review, fluckiger2007ultrasound, rubin2006sonographic, kim2008noninvasive, yim20133} were proposed to calculate the 3D position and orientation of the endoscopic capsule robot such as fluoroscopy \citep{than2012review}, ultrasonic imaging \citep{fluckiger2007ultrasound, rubin2006sonographic, kim2008noninvasive, yim20133}, positron emission tomography (PET) \citep{than2012review, yim20133}, magnetic resonance imaging (MRI) \citep{than2012review}, radio transmitter based techniques and magnetic field based techniques. The common drawback of these localization methods is that they require extra sensors and hardware design. Such extra sensors have their own drawbacks and limitations if it comes to their application in small scale medical devices such as space limitations, cost aspects, design incompatibilities, biocompatibility issue and the interference of the sensors with the activation system of the device. 

As a solution of these issues, a trend of vision-based localization methods have attracted the attention for the localization of such small scale medical devices. As a first attempt, structure from motion (SfM) methods have been proposed to deal with monocular endoscope localization \citep{sitti2015biomedical, goenka2014capsule}. However, SfM methods are incapable of real time processing which makes them unsuitable for robotic localization. The use of visual simultaneous localization and mapping (VSLAM) in medical field has been researched by \citep{mountney2006simultaneous, nakamura2008capsule}, who modified the extended Kalman filter SLAM (EKF-SLAM) framework from \citep{munoz2014review} for medical applications. \citep{lin2013simultaneous, liao2010indications} modified the breakthrough VSLAM method parallel tracking and mapping (PTAM) algorithm to a stereo-endoscope to build a denser 3D map than previous EKF based SLAM systems. Due to the non-rigid deformations inside human body, the use of only a monocular endoscope has remained as a challenge. \citep{grasa2009ekf, mahoney2013managing} provided extensive validation on in-vivo human sequences, e.g. demonstrating the efficiency of EKF-SLAM for hernia defect measurements in hernia repair surgery. Following the trend of PTAM modification, another milestone VSLAM method, ORB-SLAM was adjusted into medical field and proposed as ORB SLAM-based Endoscope Tracking and 3D Reconstruction method \citep{mahmoud2016orbslam} for endoscopic localization.

All of the presented vision-based localization methods use landmark-based methods, where the underlying idea is to search for distinct landmark positions and track them across frames for localization. These methods show poor performance on endoscopic images mainly because such images lack distinct features that algorithms need to extract and track to perform a precise visual odometry. Another issue in front of mono camera based VSLAM algorithms is that the translation in z direction is mathematically not possible to determine up to a scale using only one camera. Moreover, all of the existing methods in literature are proposed for handheld standard endoscopic cameras. However, endoscopic capsule robot differs from standard handheld endoscope due to its own characteristic problems such as space availability, limited energy source, low camera quality and resolution. Convolutional neural networks (CNN), such as deep neural networks, convolutional deep neural networks, deep belief networks and recurrent neural networks have been applied to fields such as computer vision, automatic speech recognition, natural language processing and bioinformatics where they have produced state-of-the-art results on various tasks and outperformed already existing  methods in these fields. There has been especially a wide-spread adoption of various deep-neural network architectures for computer vision related tasks because of the apparent empirical success these architectures have shown in various image classification tasks. In fact, the seminal paper ImageNet Classiﬁcation with Deep Convolutional Neural Networks has been cited over 8000 times \citep{krizhevsky2012imagenet}. With this intention, we propose in this paper a new method for 6-DoF localization of both endoscopic capsule robot and handheld standard endoscope using the CNN system. We demonstrate that CNN learns the most relevant feature vector representation related to 3D position and orientation estimation of the robot from 2D raw image inputs. The method we introduced solves several issues confronted with the challenge endoscopic image datasets faced by typical SLAM pipelines, such as the need to establish frame-to-frame feature correspondence which is practically not working on low textured inner organ tissues and the need to extract large baseline key frames. The main benefit of our deep learning based approach is that it does not require any additional sensor. And compared to vision based unsupervised localization methods, our supervised method making use of pre-trained information is more appropriate for endoscopic type of images since the operation environment remains more or less similar among different patients. For the first time in literature, we propose a deep learning based real time 6-DoF localization method for mono camera based endoscopic capsule robot and handheld standard endoscope. As the outline of this paper, section \ref{sec:alg_analysis} introduces the proposed CNN based localization method in detail. Section \ref{sec:dataset_equip}  presents our dataset and the experimental setup. Section \ref{sec:experiements} shows our experimental results for 6-DoF localization of the endoscopic capsule robot. Section \ref{sec:conclusion} discusses the drawbacks of our method and gives future directions.

\section{System Overview and Analysis} \label{sec:alg_analysis}

\subsection{Architecture}
This paper addresses the vision-based localization problem of the endoscopic capsule robots. Localization capability is a significant functionality for mobile robots. We introduce a novel localization method using a CNN architecture that is trained end-to-end to regress the robot's orientation and position in real time (5 ms per frame). The proposed system takes a single endoscopic RGB image and regresses the camera's 6-DoF pose without a need of any extra sensor. A loss function is minimized using the back-propagation algorithm to optimize the parameters of the architecture.

\begin{figure}
  \includegraphics[width=\textwidth]{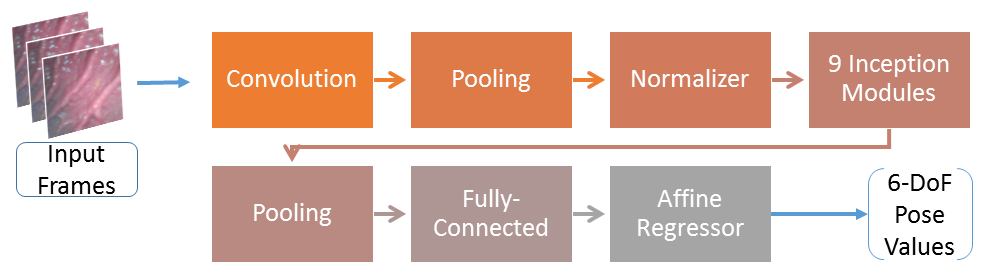}
\caption{Architecture Block Diagram.}
\label{fig:architecture}       
\end{figure}

\begin{figure}
\includegraphics[width=\textwidth]{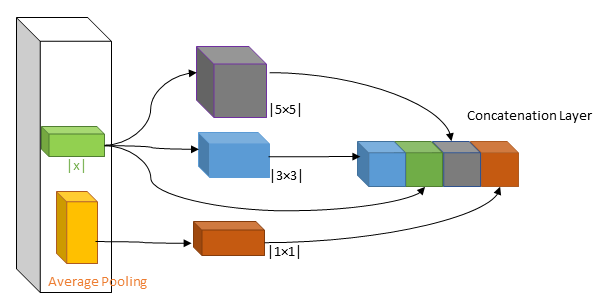}
\caption{Inception module.}
\label{fig:inception}       
\end{figure}

Inspired from GoogLeNet \citep{szegedy2015going}, the proposed architecture consists of a stack of convolutional layers, max pooling layer, normalization layer, inception modules, pooling layer, fully-connected layer and affine regressor layer, a total of 23 layers. A general block diagram of the modified network architecture can be seen in Figure \ref{fig:architecture}. We removed softmax layer from the GoogLeNet and integrated a fully-connected (FC) layer and an affine regressor layer to regress 6-DoF pose values. The hyperparameters of convolution layer such as kernel size, stride and number of filters are optimized via the training to get the most relevant information from the raw image for 6-DoF localization. Rectified-linear unit (RELU) layer, an activation unit of lower computational complexity, is robust to vanishing gradients during the training to ensure that the corresponding neural unit remains active and continues updating the learning weights. Pooling layer reduces spatial dimensions in order to lower the overfitting risk and gain computational performance. Fully connected layer connects each neuron to all activation units in the previous layer. Introduced by GoogLeNet, inception modules are the most important layers of the proposed CNN architecture increasing the learning capacity of the network. A block diagram for inception module is shown in Figure \ref{fig:inception}. Its task is the extraction of the large scale information using larger kernels while keeping to gather detail information using smaller kernels. The module basically acts as multiple convolution filter inputs, that are processed on the same input. It also does pooling at the same time. All the results are then concatenated. This allows the model to take advantage of multi-level feature extraction from each input.

For the implementation of the proposed architecture, Caffe library\citep{jia2014caffe} was used which provides a modifiable framework, enables multi-GPU training and accelerates the computational procedure by CUDA supported GPU computation. The learning rate was initialized to $0.001$ reducing as the epochs of the training continues. Adaptive Moment Estimation (Adam) \citep{kingma2014adam} optimization method was used to optimize the goal function. We run our algorithm on an Amazon EC2 p2.xlarge GPU compute instance. The list of the parameters is as follow:
\begin{itemize}
\item learning rate: $0.001$
\item momentum1: $0.9$
\item momentum2: $0.999$
\item epsilon: $10^{-8}$
\item solver type: Adam
\item batch size: $64$
\item GPU: NVIDIA K80
\end{itemize}

\begin{figure}
  \includegraphics[width=1\textwidth]{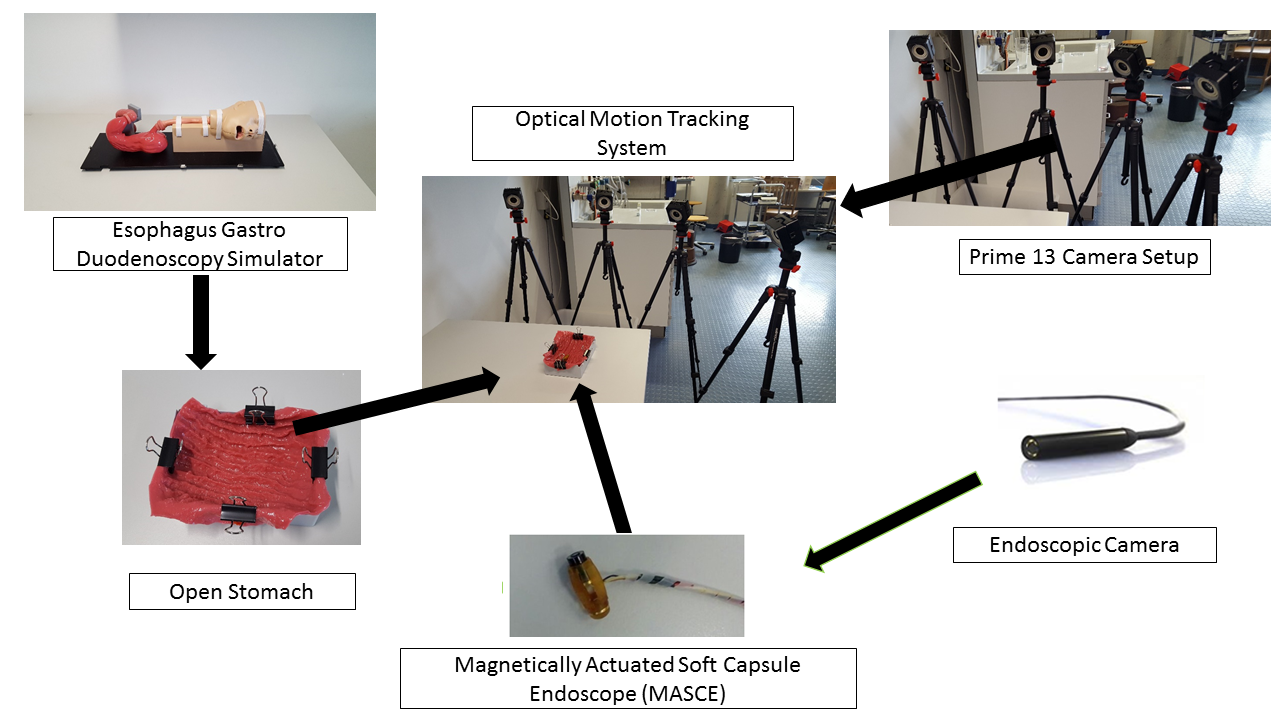}
\caption{Experimental Setup}
\label{fig:exp_setup}       
\end{figure}

\subsection{CNN-based 6-DoF Localization}
Our proposed CNN system learns translational and rotational movements simultaneously. To regress pose, we trained the CNN architecture on Euclidean loss using Adam with the following objective loss function
\begin{equation} \label{eqn:loss}
loss(I) = \|\hat{x} - x\|_2 + \beta\|\hat{q} - q\|_2
\end{equation}
where $x$ is the translation vector and $q$ is the rotation vector. The pseudocode to calculate the loss value is given in Algorithm \ref{algo:calc_loss}. In our loss function, a balance $\beta$ must be kept between the orientation and translation loss values which are highly coupled as they are learned from the same model weights \citep{kendall2015posenet}. Experimental results showed that the optimal $\beta$ was given by the ratio between expected error of position and orientation at the end of the training session.

\begin{algorithm}
\caption{Pseudo code to calculate the loss over the network}
\label{algo:calc_loss}
\begin{algorithmic}[1]
	\Procedure{CalculateLoss}{}
	\State $loss$ $\gets 0$
	\For{$layer$ in $layers$}
	\For{$top, loss\_weight$ in $layer.tops, layer.loss\_weights$}
		\State $loss \gets loss+loss\_weight \times sum(top)$
	\EndFor
	\EndFor
	\EndProcedure
\end{algorithmic}
\end{algorithm}

The back-propagation algorithm is used to calculate the gradients of CNN weights. These gradients are passed into the Adam optimization method which computes adaptive learning rates for each parameter employing the first-order gradient-based optimization of the stochastic objective function. In addition to saving exponentially decaying average of past squared gradients, $v_t$, Adam keeps exponentially decaying average of past gradients, $m_t$ that is similar to momentum.  The update equations are given as 
\begin{equation} \label{eqn:weight_update1}
(m_t)_i = \beta_1 (m_{t-1})_i + (1-\beta_1 )(\nabla L(W_t))_{i}
\end{equation}
\begin{equation} \label{eqn:weight_update2}
(v_t)_i = \beta_2 (v_{t-1})_i + (1-\beta_2)(\nabla L(W_t))_i^2
\end{equation}
\begin{equation} \label{eqn:weight_update3}
(W_{t+1})_i = (W_t)_i - \alpha \frac{\sqrt{1-(\beta_2)_i^t}}{1- (\beta_1)_i^t} \frac{(m_t)_i}{\sqrt{(v_t)_i+\varepsilon}}
\end{equation}

We used default values proposed  by \citep{kingma2014adam} for the parameters $\beta_1,\beta_1$ and $\varepsilon$: $\beta_1=0.9$, $\beta_2=0.999$ and $\varepsilon=10^{-8}$.

\subsection{Transfer Learning}
We sidestep the common problem of CNN that it requires large amount of training data by using transfer learning for the first layers of the proposed architecture. For that aim, we started the pose training using the weights acquired by ImageNet \citep{wugpu}. According to the experimental results, we were able to conclude that a network trained to output pose-invariant classification labels is suitable as a starting point for a pose regressor in the context of endoscopic capsule robot localization. A possible explanation of that observation is that the network must have activation units that are invariant to different pose values.

During transfer learning on the first layers, we kept the learning rate low in order not to distort the already learned weights. During the training of the next layers, the learning rate of individual models were increased to enable a learning procedure from scratch. We aim to publish the weights of the trained model for the sake of other research groups working on endoscopic capsule robot localization.

\begin{figure}
  \includegraphics[width=\textwidth]{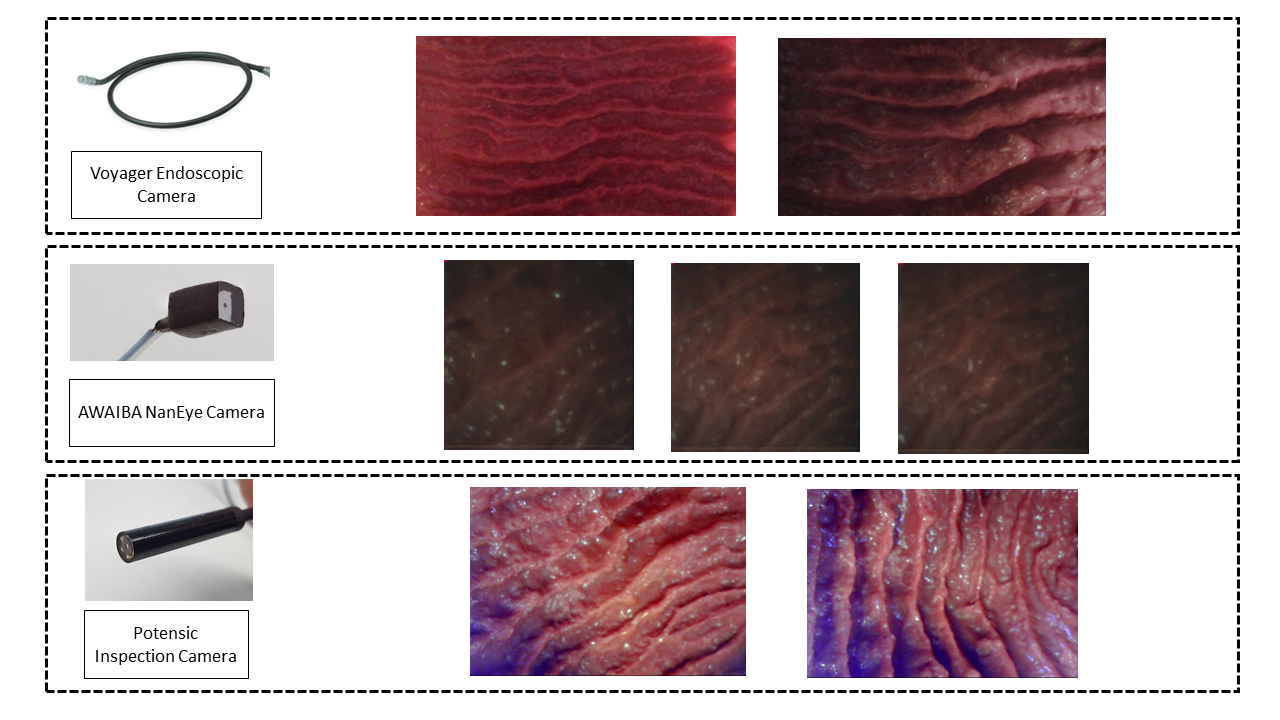}
\caption{Example images obtained from different cameras.}
\label{fig:dataset}       
\end{figure}

\section{DATASET, EQUIPMENT AND SPECIFICATIONS} \label{sec:dataset_equip}
\begin{figure}
\centering
  \includegraphics[width=0.5\textwidth]{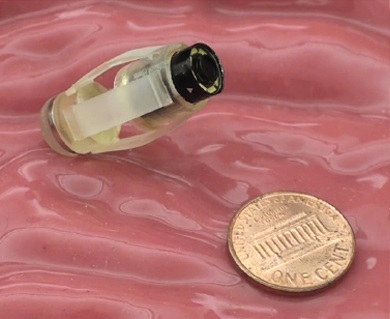}
\caption{Photo of the endoscopic capsule robot prototype used in the experiments.}
\label{fig:masce}       
\end{figure}

In deep learning based applications, the existence of a large amounts of training data is  very important. Since there is no publicly available dataset existing for 6-DoF endoscopic capsule robot localization to our knowledge, we created our own dataset and labels for that purpose using different endoscopic cameras mounted on our magnetically activated soft capsule endoscope  (MASCE) system as seen in Figure \ref{fig:masce}.

\begin{figure}
\includegraphics[width=1\textwidth]{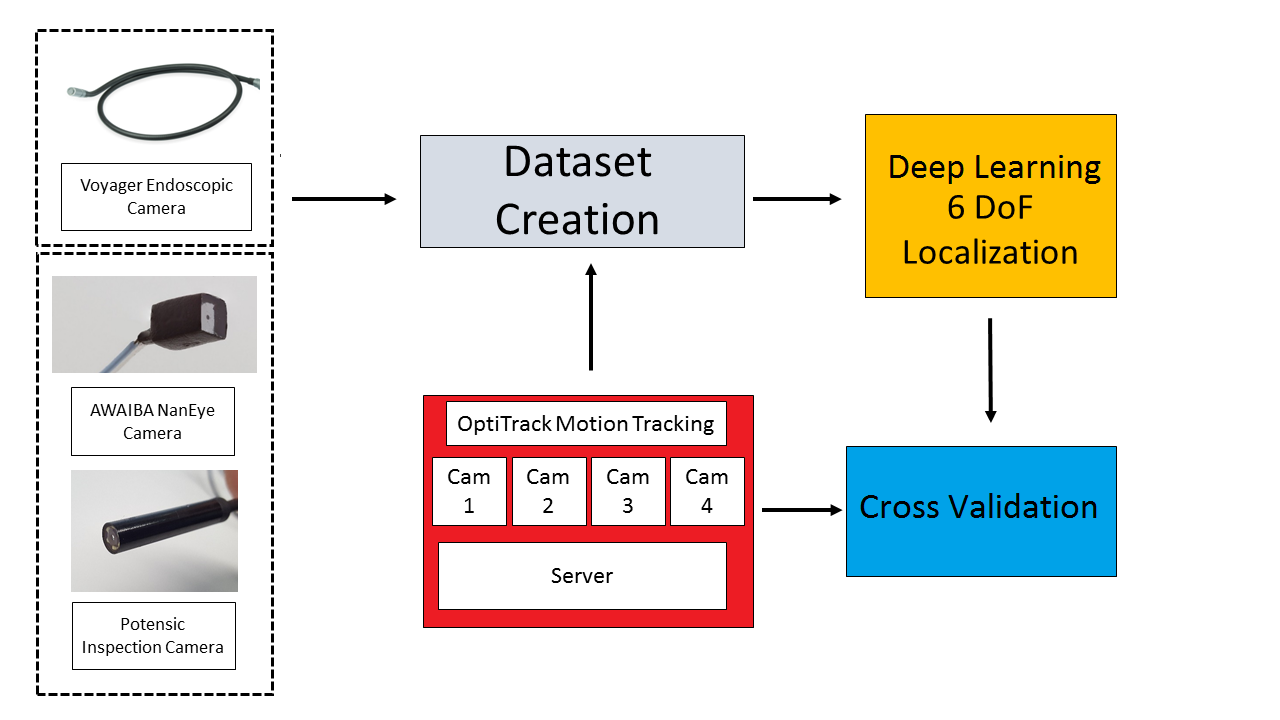}
\caption{System diagram}
\label{fig:sys_diag}       
\end{figure}

For the labelling of the endoscopic videos, OptiTrack motion tracking system consisting of four infrared cameras and a tracking software was employed (see Figure \ref{fig:sys_diag}). To make sure that our dataset was not narrowed down to just one specific endoscopic camera, three different endoscopic cameras were used to capture the endoscopic videos. The cameras used for recording the dataset were the AWAIBA Naneye, VOYAGER and POTENSIC endoscopic camera. Specifications of the cameras can be found in Table \ref{tab:1}, Table \ref{tab:2} and Table  \ref{tab:3}, respectively.

The videos were recorded on a non-rigid, realistic surgical stomach model Koken LM103 - EDG (EsophagoGastroDuodenoscopy) Simulator. We used paraffin oil to increase the reflection inside the simulator since paraffin oil has the same viscosity as the stomach fluid. For visual reference see Figure \ref{fig:exp_setup}. A total of 15 minutes of stomach imagery was recorded for this research containing over 10000 frames. Some sample frames of the dataset are shown in Figure \ref{fig:dataset}.

To avoid overfitting of the proposed CNN architecture, we extended the dataset applying different image distortion techniques on to the raw images such as Gaussian blur, median blur, brightness distortion.

\begin{table}
\caption{AWAIBA NANEYE MONOCULAR CAMERA SPECIFICATIONS}
\label{tab:1}       
\begin{tabular}{ll}
\hline\noalign{\smallskip}
Resolution & 250 x 250 pixels \\
\hline\noalign{\smallskip}
Footprint & 2.2 x 1.0 x 1.7 mm \\
\hline\noalign{\smallskip}
Pixel size & 3 x 3 $\mu m^2$\\
\hline\noalign{\smallskip}
Pixel depth & 10 bit\\
\hline\noalign{\smallskip}
Frame rate & 44 fps\\
\hline\noalign{\smallskip}
\end{tabular}
\end{table}

\begin{table}
\caption{POTENSIC MINI MONOCULAR CAMERA SPECIFICATIONS}
\label{tab:2}       
\begin{tabular}{ll}
\hline\noalign{\smallskip}
Resolution & 1280 x 720 pixels \\
\hline\noalign{\smallskip}
Footprint & 5.2 x 4.0 x 2.7 mm \\
\hline\noalign{\smallskip}
Pixel size & 10 x 10 $\mu m^2$\\
\hline\noalign{\smallskip}
Pixel depth & 10 bit\\
\hline\noalign{\smallskip}
Frame rate & 15 fps\\
\hline\noalign{\smallskip}
\end{tabular}
\end{table}

\begin{table}
\caption{VOYAGER MINI CAMERA SPECIFICATIONS}
\label{tab:3}       
\begin{tabular}{ll}
\hline\noalign{\smallskip}
Resolution & 720 x 480 pixels \\
\hline\noalign{\smallskip}
Footprint & 5.2 x 5.0 x 2.7 mm \\
\hline\noalign{\smallskip}
Pixel size & 10 x 10 $\mu m^2$\\
\hline\noalign{\smallskip}
Pixel depth & 10 bit\\
\hline\noalign{\smallskip}
Frame rate & 15 fps\\
\hline\noalign{\smallskip}
\end{tabular}
\end{table}

\begin{figure}[t!]
\begin{subfigure}[t]{0.5\textwidth}
\includegraphics[width=\textwidth]{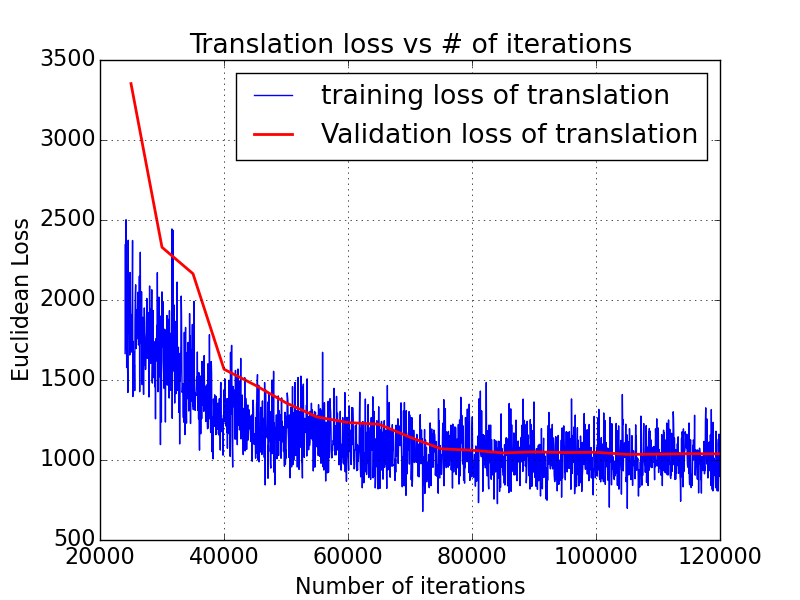}
\label{fig:trans_err}       
\end{subfigure}%
    ~ 
\begin{subfigure}[t]{0.5\textwidth}
\includegraphics[width=\textwidth]{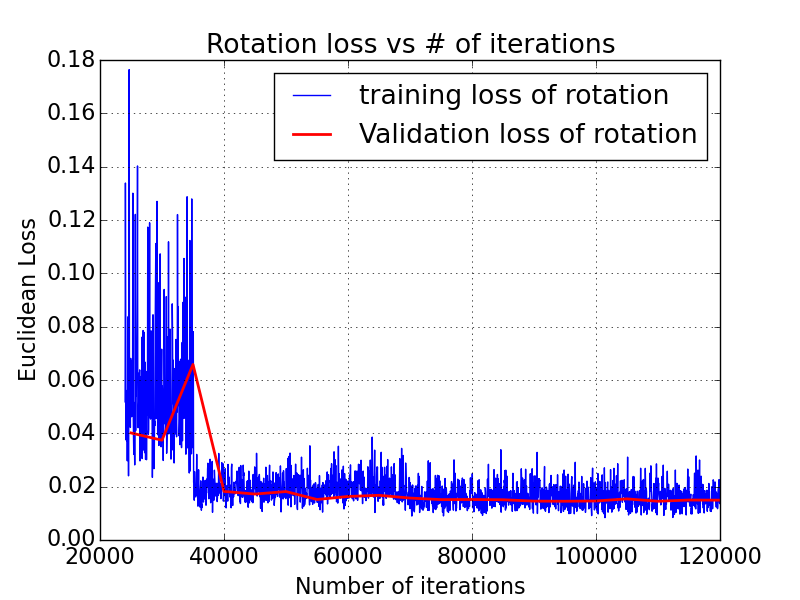}
\label{fig:rot_err}       
\end{subfigure}
\caption{The convergence of the loss for validation and training datasets. Both rotation and translation errors of the validation and the training datasets converge.}
\label{fig:loss_it}
\end{figure}

\section{EXPERIMENTS AND RESULTS} \label{sec:experiements}

\begin{figure}[t!]
\begin{subfigure}[t]{0.5\textwidth} 
\includegraphics[width=\textwidth]{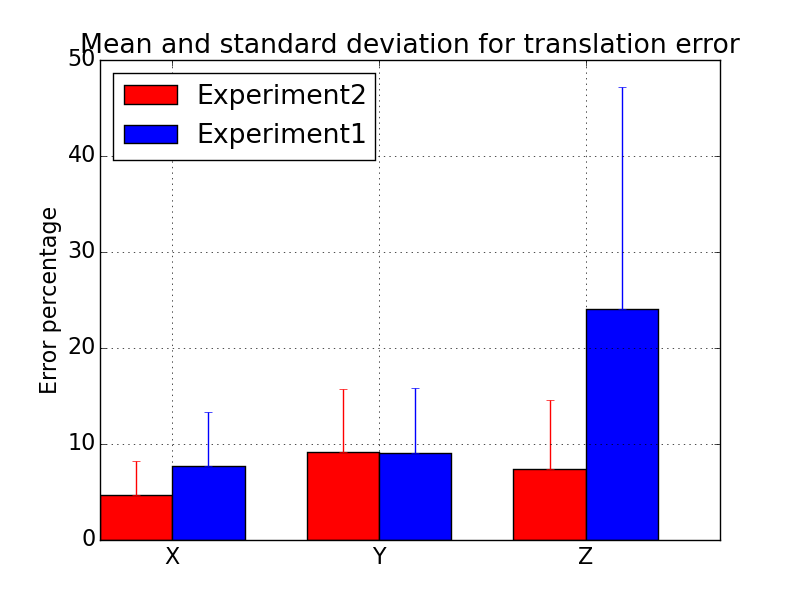}
\label{fig:cross_trans}       
\end{subfigure}
~
\begin{subfigure}[t]{0.5\textwidth} 
\includegraphics[width=\textwidth]{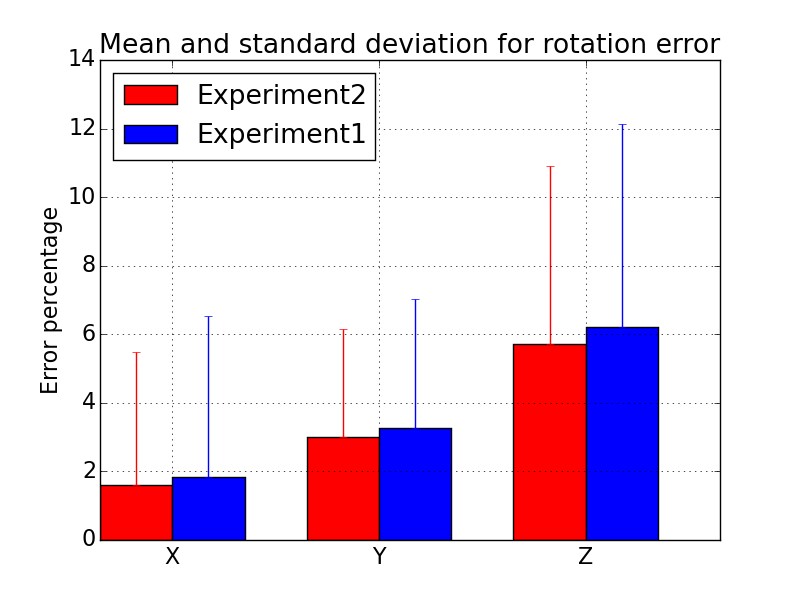}
\label{fig:cross_rot} 
\end{subfigure}
\caption{Cross validation results for translation rotation errors.}
\label{fig:cross_val}
\end{figure}

Cross validation method was used  to assess the performance of the proposed CNN based localization method. The dataset was divided into two parts of which the first part was used for training purposes and the second part for validation purposes. For a quantitative analysis, we compared the estimated 6-DoF pose values with the ground truth data. The mean and standard deviation of the translational and rotational errors in x, y and z axis are shown in Figure \ref{fig:cross_val}.

\begin{figure}[t!]
\begin{subfigure}[t]{\textwidth} 			
\includegraphics[width=\textwidth]{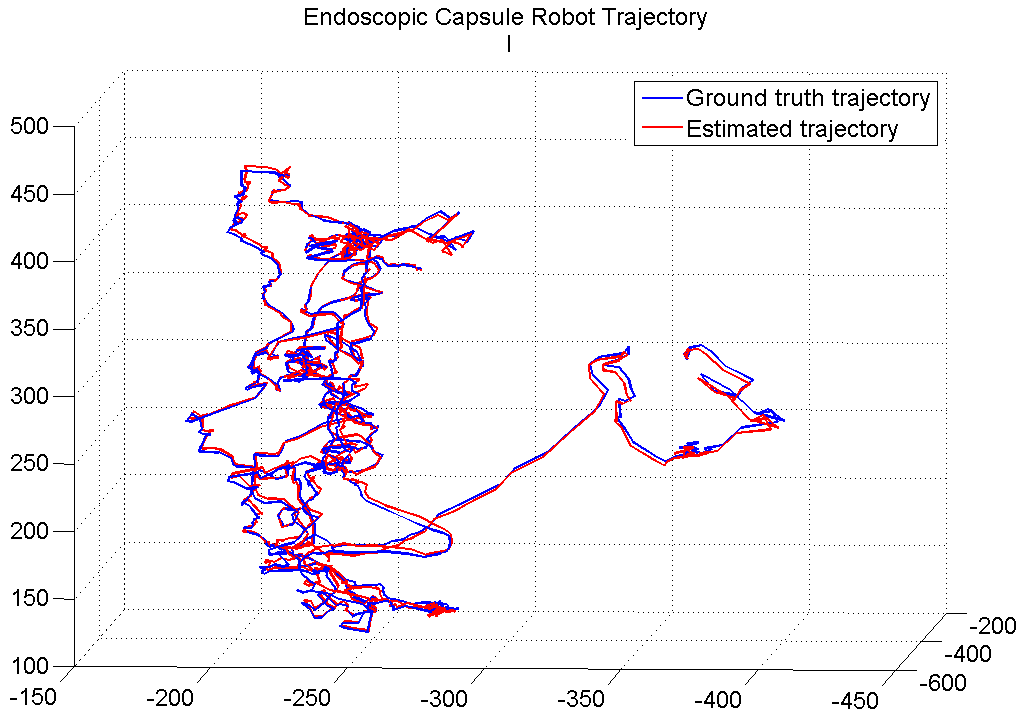}
\end{subfigure}%
   \\
\begin{subfigure}[t]{\textwidth}
\includegraphics[width=\textwidth]{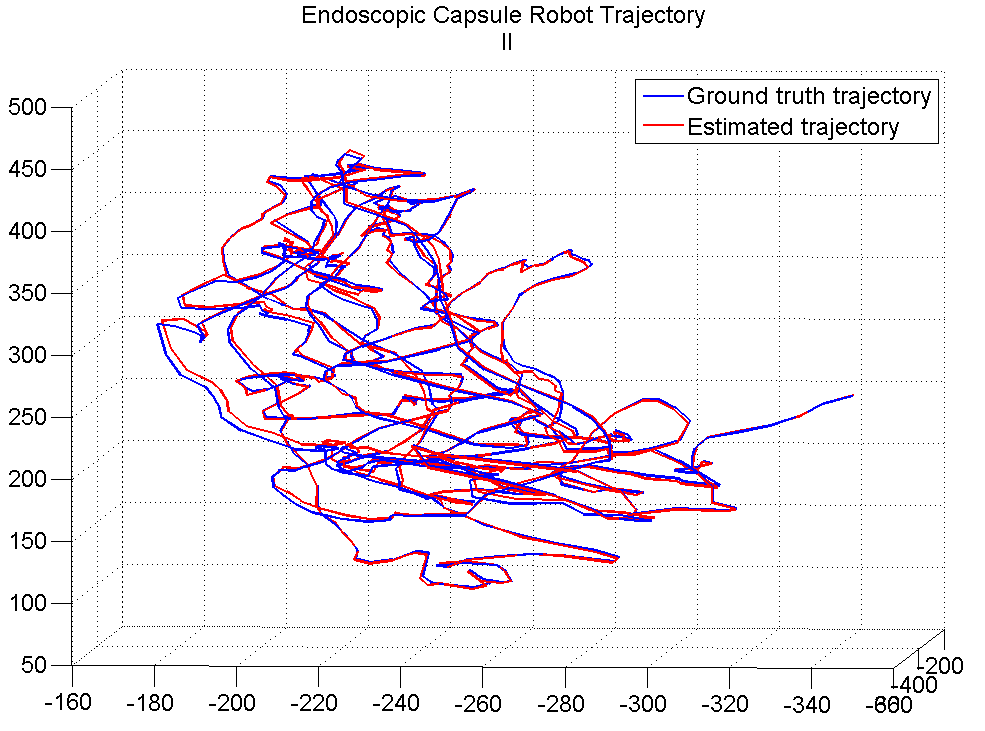}
\end{subfigure}%
    ~ 
\caption{Estimated and ground truth 3D trajectories  for two different cases.}
\label{fig:3d_traj}       
\end{figure}

We conducted two separate experiments for the training of the CNN network. The first experiment was conducted with 10000 frames of which approximately 7000 frames were used for training and 3000 frames for validation. In the second experiment, 70000 frames were used partially consisting of the raw images and partially consisting of the distorted data we generated from the original dataset. We run both experiments on an Amazon EC2 p2.xlarge machine instance. The duration of the first training session was 4 hours and 37 minutes and the duration of the second training session was 19 hours and 17 minutes. After the training sessions, we tested the proposed CNN based localization method on a test dataset that was never seen during the training session by the CNN. The 6-DoF pose estimation per image takes 5 ms on a standard CPU Intel i5 core desktop PC.

Figure \ref{fig:loss_it} shows the training error (blue) and validation error (red) of translational and rotational movements both as a function of the number of training cycles. Both the validation and training error decrease and converge to a global minimum, where the learning cycle was stopped to avoid any overfitting and underfitting of the dataset.

Better results were obtained in the second experiment: The best rotation accuracy is achieved with respect to x axis with an error of $1.60\%$ whereas the second rotational accuracy is achieved with respect to y axis with an error of $3.01\%$. The rotational error around the z axis is $5.71\%$. The translation errors are $4.72\%$, $9.16\%$ and $7.44\%$ in x, y and z axis, respectively. Figure \ref{fig:3d_traj} represents for two different challenge cases consisting of fast rotations and large translations the 3D trajectory plots acquired by the OptiTrack system (blue) and the 3D trajectory estimated by our method (red). As observed in both cases, the proposed CNN system is able to track the endoscopic capsule robot motion very accurately.

\begin{figure}
\includegraphics[width=1\textwidth]{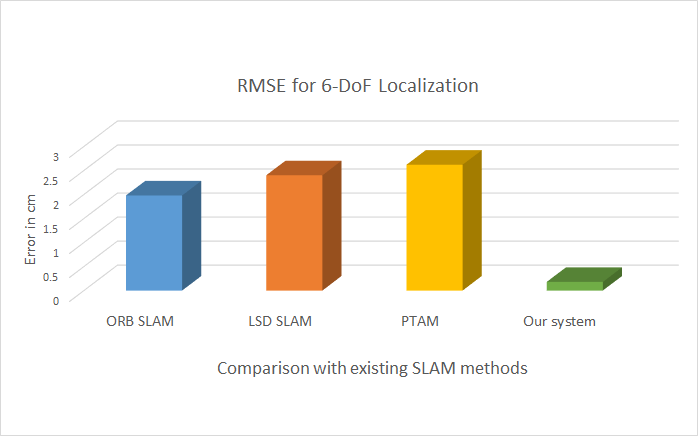}
\caption{Comparison of the proposed method with existing SLAM methods.}
\label{fig:methods_comp}       
\end{figure}

Figure \ref{fig:methods_comp} compares our proposed system with the existing popular SLAM methods including ORB SLAM, PTAM and LSD SLAM. To compare the algorithms, we calculated the root-mean-square error (RMSE) of the estimated results with respect to the OptiTrack ground truth. As shown in the Figure \ref{fig:methods_comp}, PTAM shows the worst performance out of all methods with an RMSE of $2.6$ cm while our proposed system outperforms all of them with an RMSE of $0.18$ cm for a trajectory of $18$ cm.

\section{CONCLUSION} \label{sec:conclusion}
We present, to our knowledge, the first application of
deep CNN to end-to-end 6-DoF localization for endoscopic capsule robots and handheld standard endoscopes. We have demonstrated that one can overcome the need for huge amount of training data for CNN with the help of transfer learning. We also concluded that the pre-trained weights on ImageNet dataset indeed reveals information about pose values and they are useful for stomach environment as well even though it is not directly related to the endoscopic environment and robot localization. Our method tolerates many issues faced by modern SLAM techniques such as feature correspondence establishment across frames in low textured areas, high reflections, motion blur and low image quality. Since huge amount of training dataset is not easy to create for endoscopic applications, we consider to publish our trained weights for 6-DoF endoscopic capsule robot localization online for the sake of other groups working in that area.

As future work, we aim to use our trained deep learning architecture as a deep feature extractor and serve such deep features as an input to modern SLAM methods which will eliminate the need for long training sessions. Additionally, we consider to extend the proposed deep learning based localization method to stereo endoscopic capsule robot case.


\section*{References}

\bibliography{mybibfile}

\end{document}